\documentclass{article}

\usepackage{PRIMEarxiv}

\usepackage[utf8]{inputenc}
\usepackage[T1]{fontenc}
\usepackage{hyperref}
\usepackage{url}
\usepackage[numbers]{natbib}
\usepackage{booktabs}
\usepackage{amsfonts}
\usepackage{amssymb}
\usepackage{nicefrac}
\usepackage{microtype}
\usepackage{graphicx}
\usepackage{subcaption}
\usepackage{float}
\usepackage{amsthm}
\usepackage[capitalize,noabbrev]{cleveref}

\usepackage{amsmath,amsfonts,bm}

\def\eqref#1{equation~\ref{#1}}

\def\1{\bm{1}}

\DeclareMathAlphabet{\mathsfit}{\encodingdefault}{\sfdefault}{m}{sl}
\SetMathAlphabet{\mathsfit}{bold}{\encodingdefault}{\sfdefault}{bx}{n}

\theoremstyle{plain}

\theoremstyle{definition}

\theoremstyle{remark}

\title{Beyond Uniform Credit: Causal Credit Assignment for Policy Optimization}

\author{
  Mykola Khandoga, Rui Yuan, Vinay Kumar Sankarapu \\
  \affiliation{Lexsi Labs}
}

\setabstract{%
Policy gradient methods for language model reasoning, such as GRPO and DAPO, assign uniform credit to all generated tokens - the filler phrase ``Let me think'' receives the same gradient update as the critical calculation ``$23 + 45 = 68$.'' We propose counterfactual importance weighting: mask reasoning spans, measure the drop in answer probability, and upweight tokens accordingly during policy gradient updates. Our method requires no auxiliary models or external annotation, instead importance is estimated directly from the policy model's own probability shifts. Experiments on GSM8K across three models spanning the Qwen and Llama families demonstrate consistent improvements over uniform baselines and faster convergence to equivalent accuracy. Inverting the importance signal hurts performance, confirming we capture genuine causal structure rather than noise. Analysis shows the method correctly prioritizes calculation steps over scaffolding text. We view these findings as establishing counterfactual importance weighting as a foundation for further research rather than a complete solution.
}

\setkeywords{Reinforcement Learning, Language Models, Credit Assignment, Policy Optimization}

\runningtitle{Beyond Uniform Credit: Causal Credit Assignment for Policy Optimization}

\begin{document}
\maketitle

\section{Introduction}

The advancements in reinforcement learning with verifiable rewards (RLVR) and the GRPO algorithm \citep{shao2024deepseekmath} along with its successors - DAPO \citep{yu2025dapo}, Dr. GRPO \citep{liu2025drgrpo}, and others - have enabled rapid improvements in model capabilities on tasks like mathematics and coding. When GRPO proposed eliminating the critic model, it led to substantial savings in computation and VRAM. But this simplification came at a cost: the granularity of credit assignment was reduced. Now all tokens in a generated solution receive exactly the same reward signal, regardless of their actual contribution to the final answer.

There have been attempts to address this problem through various means. Process Reward Models (PRMs) \citep{lightman2023lets, wang2024mathshepherd} provide step-level supervision by training a separate model to score intermediate reasoning steps - but this requires either expensive human annotation or synthetic labels from a stronger model. Token-level reward methods like TLCR \citep{yoon2024tlcr} train discriminators to assign per-token rewards, adding architectural complexity. Segment-based approaches like SPO \citep{guo2025spo} partition sequences at fixed intervals, but this does not adapt to the actual structure of reasoning.

We propose to use the model's intrinsic signals to determine which parts of the reasoning trace matter most for achieving the correct answer. This is achieved through counterfactual importance: we mask candidate reasoning spans and measure the resulting drop in answer probability. Spans whose removal damages the answer are deemed important and receive higher weight during training. The method requires no auxiliary models and no external annotation. We describe the approach in detail in \cref{sec:method}.

\textbf{Research Question:} Can we improve GRPO training by assigning higher credit to tokens that causally contribute to correct answers?

We address this question by introducing \textbf{counterfactual importance weighting} for GRPO-like training algorithms. Our method identifies reasoning spans (arithmetic expressions, intermediate calculations) and measures their causal importance by masking each span and observing the drop in answer probability. Spans whose removal significantly damages the answer receive higher importance scores, which then weight the corresponding tokens during gradient computation.

We focus on mathematical reasoning, where task success is determined by a single verifiable answer. This creates a natural `answer sink' --- a concentrated verification signal that makes counterfactual importance estimation well-defined: masking a critical calculation step substantially damages the probability of reaching the correct answer. We hypothesize that domains with more distributed correctness criteria (e.g., code generation, where multiple structural elements must simultaneously be correct) may benefit less from span-level interventions. We test this hypothesis in Section 5.4.

To validate that our importance scores capture meaningful signal rather than noise, we compare four weighting strategies:
\begin{itemize}
    \item \textbf{Counterfactual:} Weight tokens by their measured causal importance
    \item \textbf{Inverted:} Weight tokens inversely to importance (control)
    \item \textbf{Random:} Assign random weights (control)
    \item \textbf{Uniform:} Standard DAPO with equal weights (baseline)
\end{itemize}

If counterfactual importance captures genuine signal, we expect: Counterfactual $>$ Uniform $=>$ Random $>$ Inverted. Our experiments generally confirm this ordering across multiple models.

\paragraph{Contributions.}
\begin{enumerate}
    \item We propose a counterfactual method for measuring token-level importance in reasoning chains, based on answer probability drop under span masking.
    \item We integrate this importance signal into DAPO training through multiplicative token weighting.
    \item We provide systematic experiments across model scales (1.7B--3B parameters) showing consistent improvements of \textbf{0.8--1.1} percentage points over baseline DAPO on GSM8K.
    \item We validate the importance signal through ablations: inverted weighting consistently hurts performance, confirming the method captures genuine causal structure.
    \item We provide a detailed empirical analysis of what makes reasoning spans important: calculation chains are 11$\times$ enriched in critical spans, 3.5\% of spans are distractors that hurt performance, and counterfactual weighting concentrates 1.6$\times$ more gradient mass on high-importance tokens. These findings characterize the structure of reasoning traces and may inform future credit assignment methods.
\end{enumerate}

We view this work as a proof of concept: demonstrating that counterfactual importance provides consistent, directional signal for credit assignment. The modest but reliable gains (0.8--1.1 pp) and strong ablation results suggest this is a promising direction, with room for future work on stronger interventions, better span detection, and combination with complementary methods.

\section{Related Work}
\label{sec:related}

Our approach builds on a rich lineage of counterfactual methods in deep learning, which we briefly trace to motivate our contribution.

\paragraph{Foundations and vision origins.}
The theoretical basis for counterfactual reasoning derives from Pearl's causal hierarchy \citep{pearl2009causality} and Rubin's potential outcomes framework \citep{rubin1974estimating}, which formalize the distinction between observational correlation and interventional causation. In deep learning, \citet{zeiler2014visualizing} introduced \textbf{occlusion sensitivity} - systematically masking image regions and measuring prediction changes - establishing the ``mask and measure'' paradigm for importance estimation. This directly implements Pearl's do-operator: importance is revealed not by what correlates with predictions, but by what changes them when removed.

\paragraph{Extension to NLP.}
\citet{li2016understanding} adapted occlusion analysis to text through \textbf{representation erasure}, removing words or hidden dimensions and observing output changes. They further introduced reinforcement learning to find minimal erasure sets that flip predictions. Subsequent work formalized these intuitions: Integrated Gradients \citep{sundararajan2017axiomatic} provided axiomatic foundations for attribution, while the ``Attention is not Explanation'' debate \citep{jain2019attention} clarified that attention weights reveal where models look, but only intervention reveals what causally matters.

\paragraph{Causal analysis in transformers.}
\citet{vig2020investigating} introduced \textbf{causal mediation analysis} to transformers, distinguishing whether information \emph{exists} in representations from whether it is \emph{used}. Their intervention procedure consists in isolating causal contributions via running counterfactual inputs while restoring specific components. \citet{meng2022locating} extended this to \textbf{causal tracing}, using clean/corrupted/patched forward passes to locate where factual knowledge is stored, enabling targeted model editing. These methods demonstrated that causal understanding of transformers is both achievable and actionable.

\paragraph{Counterfactual credit assignment in RL.}
Counterfactual reasoning entered reinforcement learning through COMA \citep{foerster2018counterfactual}, which uses baselines marginalizing over alternative actions to isolate each agent's causal contribution. \citet{mesnard2021counterfactual} formalized counterfactual credit assignment by conditioning value functions on future outcomes to isolate each action's causal contribution to returns. For LLMs specifically, VinePPO \citep{kazemnejad2024vineppo} demonstrated that standard value networks fail at credit assignment in reasoning tasks, proposing Monte Carlo estimation from intermediate states. Their key finding - that most reasoning tokens do not affect problem-solving probability - directly motivates identifying the few that do.

\paragraph{Token-level methods for LLM training.}
Recent work has pursued finer-grained credit assignment through various proxies: Process Reward Models \citep{lightman2023lets} provide step-level supervision but evaluate correctness rather than causal contribution; TLCR trains discriminators for token-level rewards; RTO \citep{zhong2024rto} extracts implicit token rewards from DPO probability ratios; SPO \citep{guo2025spo} partitions sequences into segments with Monte Carlo advantage estimation. SCAR \citep{cao2025scar} applies Shapley values to distribute rewards among tokens - the closest existing work to explicit counterfactual analysis.

\paragraph{The gap we address.}
Despite this convergent trajectory, no existing method directly asks: \emph{``What would the outcome be if this token were different?''} Current approaches estimate importance through learned discrimination, probability ratios, or game-theoretic values - but not through intervention. Our counterfactual importance method completes this arc: we adapt the mask-and-measure methodology from interpretability to provide \textbf{training signals} for policy optimization, using the same interventional logic that revealed which pixels matter for image classification to identify which tokens matter for reasoning.

\paragraph{Concurrent work.}
Concurrent work by \citet{ruan2025cft} proposes Critical Token Fine-Tuning (CFT), which identifies critical tokens for supervised fine-tuning by checking whether alternative token choices preserve answer correctness. Our work addresses a complementary setting: credit assignment in reinforcement learning with policy gradients. Where CFT produces a binary mask (train or skip), we compute continuous importance weights based on answer probability drop, enabling finer-grained gradient modulation. Where CFT requires full regeneration to verify correctness, our span-level masking requires only forward passes. The convergent finding - that counterfactual reasoning identifies tokens that matter - suggests this is a general principle applicable across training paradigms.

\section{Method}
\label{sec:method}

\subsection{Background: DAPO}

We build on DAPO \citep{yu2025dapo}, which extends GRPO with decoupled clipping and token-level policy gradient loss. For a prompt $x$ with $G$ sampled completions $\{y_1, \ldots, y_G\}$ and binary rewards $\{r_1, \ldots, r_G\}$, the advantage for completion $i$ is:
\begin{equation}
    A_i = \frac{r_i - \mu_r}{\sigma_r + \epsilon}
\end{equation}
where $\mu_r$ and $\sigma_r$ are the group mean and standard deviation. The policy gradient loss is:
\begin{equation}
    \mathcal{L}_{\text{DAPO}} = -\frac{1}{\sum_i T_i} \sum_{i=1}^{G} \sum_{t=1}^{T_i} A_i \cdot \log \pi_\theta(y_t^{(i)} | x, y_{<t}^{(i)})
\end{equation}

Critically, the advantage $A_i$ is uniform across all tokens in completion $i$.

We implement the DAPO algorithm using HuggingFace TRL's \texttt{loss\_type="dapo"} option, which includes token-level loss normalization and decoupled clipping (Clip-Higher) as described by \citet{yu2025dapo}. Dynamic sampling is not used. For clarity, we present the unclipped token-sum form above. Our contribution is the addition of per-token importance weights $w_t$.

\paragraph{Intuition: Addressing gradient dilution.}
In standard GRPO-like algorithms, the policy gradient averages over all tokens equally, effectively ``diluting'' the learning signal with updates from filler phrases, boilerplate setup (``Let me solve this step by step...''), and redundant restatements. Our weighting scheme concentrates gradient mass on tokens with demonstrated causal influence, namely those whose removal damages answer probability. This can be viewed as a non-parametric alternative to Process Reward Models: where PRMs require training an auxiliary value function $V(s)$ to estimate step-wise contributions \citep{lightman2023lets}, counterfactual masking provides direct estimates of causal importance using only the policy model itself, with no additional parameters or training data. The connection to \citet{mesnard2021counterfactual}'s ``skill vs.\ luck'' decomposition is direct: by isolating tokens that \emph{caused} success rather than merely co-occurred with it, we bias gradient updates toward the reasoning steps that actually determined the outcome. Our analysis confirms this intuition: calculation chains receive 11$\times$ higher importance than scaffolding tokens (\cref{tab:content_patterns}), and 3.5\% of spans are outright distractors whose removal \emph{improves} answer probability---content that uniform weighting would incorrectly reinforce.

\subsection{Counterfactual Importance Estimation}

We measure token importance through causal intervention. Let a completion $y$ be decomposed as $y = (r, a)$, where $r$ is the reasoning prefix and $a$ is the final answer span.

\paragraph{Answer-probability drop.}
For each detected reasoning span $s_k \subset r$, we define the counterfactual \emph{drop} as:
\begin{equation}
    D(s_k) = \log P_\theta(a \mid x, r_{-s_k}) - \log P_\theta(a \mid x, r)
\end{equation}
where $r_{-s_k}$ denotes the reasoning prefix with span $s_k$ replaced by a fixed placeholder string. More negative $D(s_k)$ indicates higher causal importance: masking the span damages answer probability. Positive $D(s_k)$ indicates a \emph{distractor} span whose removal actually \emph{improves} answer probability.

We convert this into a non-negative \emph{importance score} for weighting:
\begin{equation}
    I(s_k) = -D(s_k)
\end{equation}
High $I(s_k)$ indicates the span is causally necessary for the model to produce the correct answer.

\paragraph{Computing $P_\theta(a \mid x, r)$.}
We compute $\log P_\theta(a \mid x, r)$ by teacher forcing over the answer tokens:
\begin{equation}
    \log P_\theta(a \mid x, r) = \sum_{t=1}^{|a|} \log \pi_\theta(a_t \mid x, r, a_{<t})
\end{equation}
All counterfactual forward passes are run under \texttt{torch.no\_grad()} to avoid memory overhead.

\paragraph{Placeholder design.}
We use replacement rather than deletion to preserve positional structure and avoid length-change confounds. The placeholder token is the model's pad token (or end-of-text token if pad is unavailable), repeated to match the span length.

\paragraph{Why spans, not tokens?} We operate at span granularity rather than per-token for two reasons. First, computational cost: per-token masking would require 200--500 forward passes per completion (one per token), whereas span-level masking requires only 5--10, reducing overhead by 20--50$\times$. Second, semantic coherence: individual tokens lack standalone meaning - ``23'', ``+'', ``45'' separately do not constitute a reasoning step, whereas the span ``23 + 45 = 68'' represents a complete arithmetic operation.

\paragraph{Span detection.}
We identify reasoning spans via pattern matching: arithmetic expressions (e.g., \texttt{23 + 45 = 68}), intermediate calculations, and sentence boundaries. We process up to $K_{\max}=10$ spans per completion.

\paragraph{Weight assignment.}
For each token $t$ within span $s_k$, we assign importance weight proportional to $I(s_k)$. To ensure weights remain bounded, we first normalize importance scores within each completion to $[0, 1]$:
\begin{equation}
    \hat{I}(s_k) = \frac{I(s_k) - \min_j I(s_j)}{\max_j I(s_j) - \min_j I(s_j) + \epsilon}
\end{equation}
then map to the weight range $[w_{\min}, w_{\max}]$:
\begin{equation}
    w_t = w_{\min} + \hat{I}(s_k) \cdot (w_{\max} - w_{\min})
\end{equation}
Tokens outside detected spans receive baseline weight $w_t = 1$. Final answer tokens receive a fixed boost ($w_{\text{ans}} = 1.5$).

\subsection{Importance-Weighted Policy Gradient}

We modify the DAPO loss by introducing per-token importance weights:
\begin{equation}
    \mathcal{L}_{\text{CF-DAPO}} = -\frac{1}{\sum_i T_i} \sum_{i=1}^{G} \sum_{t=1}^{T_i} w_t^{(i)} \cdot A_i \cdot \log \pi_\theta(y_t^{(i)} | x, y_{<t}^{(i)})
\end{equation}

This formulation upweights gradient contributions from high-importance tokens (critical calculations) while downweighting filler tokens.

\subsection{Ablation Conditions}

To validate that our importance scores capture genuine signal, we compare four weighting modes:

\begin{table}[H]
\centering
\small
\begin{tabular}{lp{9cm}}
\toprule
\textbf{Mode} & \textbf{Weight Assignment} \\
\midrule
Counterfactual & $w_t \propto \hat{I}(s_k)$ for tokens in span $s_k$ (normalized importance) \\
Inverted & $w_t \propto (1 - \hat{I}(s_k))$  -  high-importance spans receive \emph{low} weight \\
Random & $w_t \sim \text{Uniform}(w_{\min}, w_{\max})$, independently per token \\
Vanilla & $w_t = 1$ for all $t$ (standard DAPO, uniform credit) \\
\bottomrule
\end{tabular}
\end{table}

If counterfactual importance captures meaningful signal, we expect: Counterfactual $>$ DAPO $\geq$ Random $>$ Inverted.

\section{Experiments}

\subsection{Setup}

\paragraph{Models.} We evaluate on three base models spanning different architectures and scales: Qwen3-1.7B, Qwen2.5-3B, and Llama3.2-3B. All models are trained with LoRA adapters \citep{hu2022lora} ($r=32$, $\alpha=32$) for parameter efficiency.

\paragraph{Dataset.} We train and evaluate on GSM8K \citep{cobbe2021gsm8k}, a dataset of 7,473 grade-school math problems requiring multi-step arithmetic reasoning. We evaluate on the standard 1,319-problem test set using exact-match accuracy with numeric parsing.

\paragraph{Training.} We use DAPO with $G=8$ completions per prompt, learning rate $2.5 \times 10^{-5}$, batch size $16$ with $4$ gradient accumulation steps, for $500$ gradient steps. We use temperature $0.6$ and top-$p$ $0.95$ for generation. Each configuration is run with $3$ random seeds.

\paragraph{Reward.} We use binary outcome reward: $r=1$ if the extracted numeric answer matches the gold answer, $r=0$ otherwise.

\paragraph{Weighting Hyperparameters.} For counterfactual and ablation conditions, we use weight range $[w_{\min}, w_{\max}] = [0.5, 4.0]$, answer boost $w_{\text{ans}} = 1.5$, and maximum $K_{\max}=10$ spans per completion.

\subsection{Main Results}

\begin{table}[H]
\centering
\small
\caption{GSM8K test accuracy (\%) at step 500. Mean $\pm$ std over seeds. Best in \textbf{bold}.}
\label{tab:main}
\begin{tabular}{lcccc}
\toprule
\textbf{Model} & \textbf{CF} & \textbf{DAPO} & \textbf{Random} & \textbf{Inverted} \\
\midrule
Qwen3-1.7B   & \textbf{84.3$\pm$0.5} & 83.4$\pm$0.6 & 82.5$\pm$0.5 & 81.9$\pm$0.9 \\
Qwen2.5-3B   & \textbf{86.7$\pm$0.2} & 85.6$\pm$0.2 & 85.8$\pm$0.8 & 85.3$\pm$0.3 \\
Llama3.2-3B  & \textbf{78.9$\pm$0.3} & 78.2$\pm$0.6 & 78.1$\pm$0.6 & 76.4$\pm$1.5 \\
\bottomrule
\end{tabular}
\end{table}

\begin{table}[H]
\centering
\small
\caption{GSM8K AUC (\%), measuring cumulative accuracy across training. Mean $\pm$ std over seeds. Best in \textbf{bold}.}
\label{tab:auc}
\begin{tabular}{lcccc}
\toprule
\textbf{Model} & \textbf{CF} & \textbf{DAPO} & \textbf{Random} & \textbf{Inverted} \\
\midrule
Qwen3-1.7B   & \textbf{83.2$\pm$0.4} & 82.5$\pm$0.5 & 82.1$\pm$0.5 & 81.0$\pm$0.8 \\
Qwen2.5-3B   & \textbf{85.8$\pm$0.4} & 85.3$\pm$0.4 & 85.0$\pm$0.8 & 84.8$\pm$0.3 \\
Llama3.2-3B  & \textbf{77.3$\pm$0.3} & 76.6$\pm$0.6 & 75.9$\pm$0.5 & 75.6$\pm$1.2 \\
\bottomrule
\end{tabular}
\end{table}

\begin{table}[H]
\centering
\small
\caption{Statistical significance of CF vs.\ DAPO (paired $t$-test across seeds). $^*p<0.05$.}
\label{tab:significance}
\begin{tabular}{lcccc}
\toprule
\textbf{Model} & \textbf{$\Delta$@500} & \textbf{$p$@500} & \textbf{$\Delta$AUC} & \textbf{$p$AUC} \\
\midrule
Qwen3-1.7B   & +0.90\% & 0.112 & +0.74\% & 0.035$^*$ \\
Qwen2.5-3B   & +1.09\% & 0.002$^*$ & +0.46\% & 0.049$^*$ \\
Llama3.2-3B  & +0.76\% & 0.125 & +0.46\% & 0.613 \\
\bottomrule
\end{tabular}
\end{table}

\begin{figure}[H]
    \centering
    \begin{subfigure}[b]{0.33\textwidth}
        \centering
        \includegraphics[width=\textwidth]{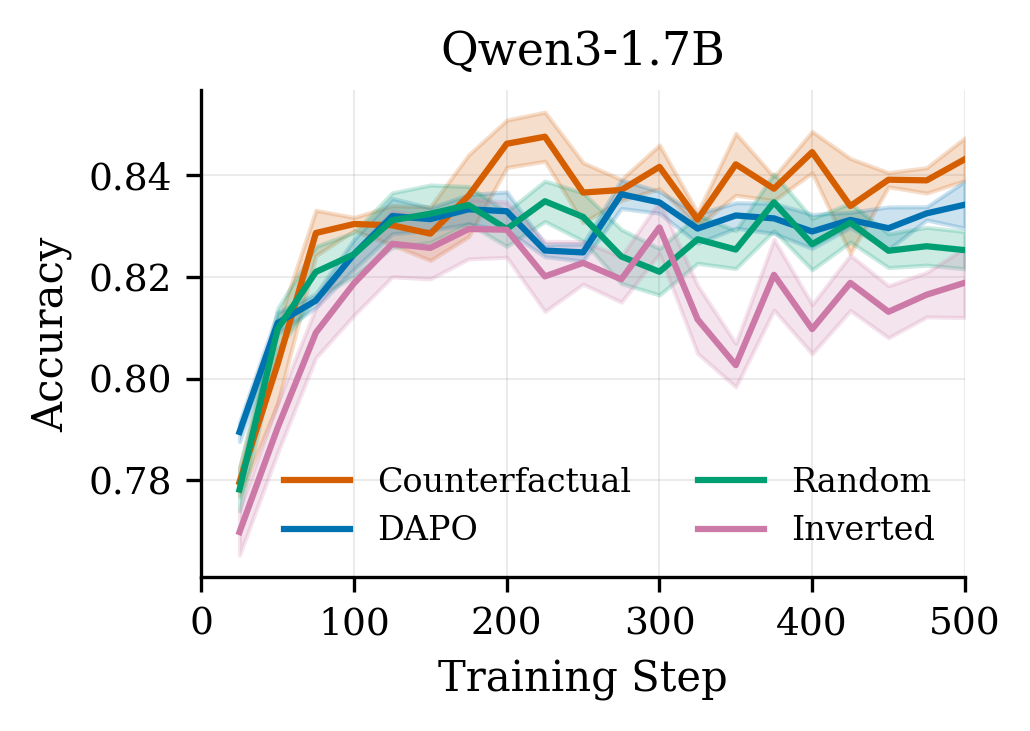}
        \caption{Qwen3-1.7B}
    \end{subfigure}
    \hfill
    \begin{subfigure}[b]{0.33\textwidth}
        \centering
        \includegraphics[width=\textwidth]{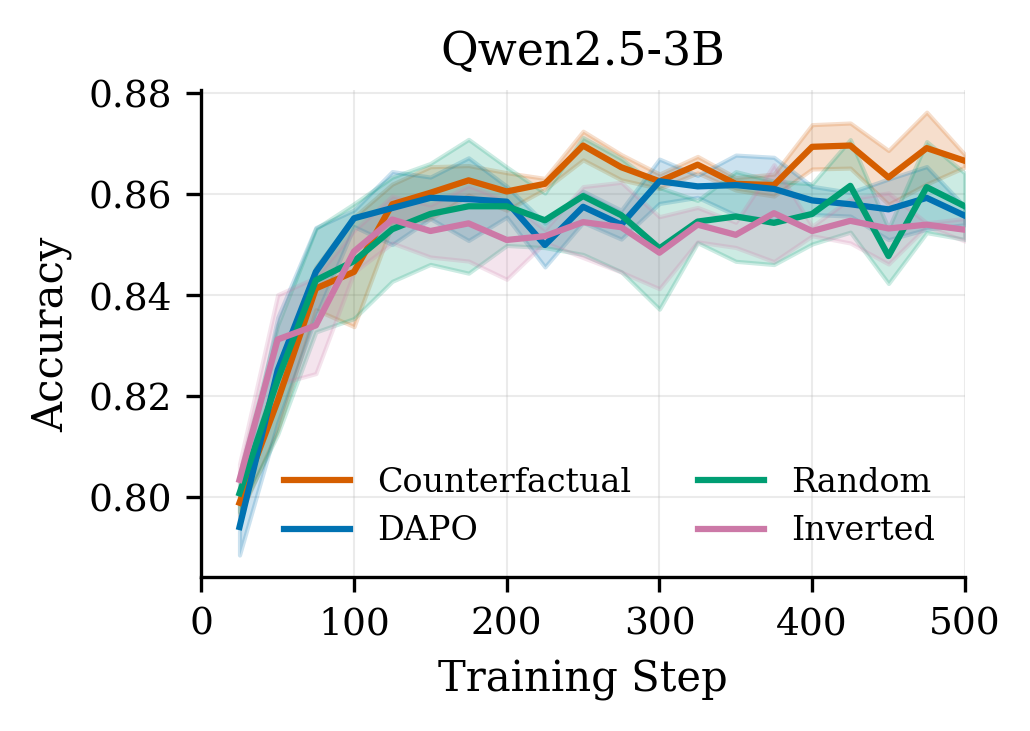}
        \caption{Qwen2.5-3B}
    \end{subfigure}
    \hfill
    \begin{subfigure}[b]{0.33\textwidth}
        \centering
        \includegraphics[width=\textwidth]{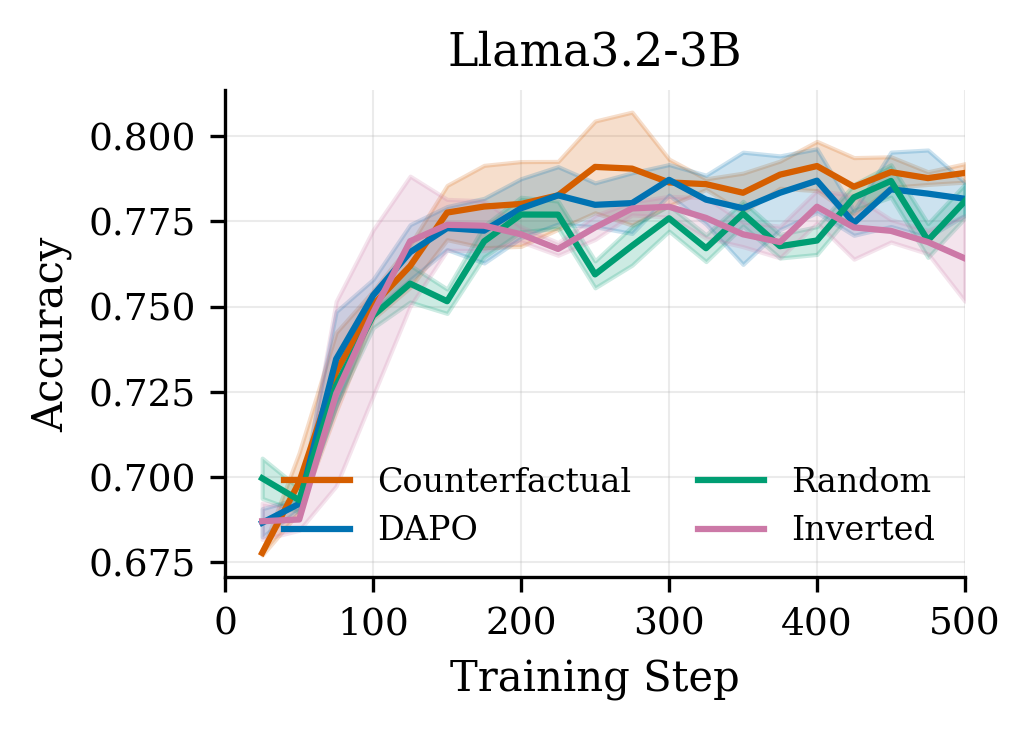}
        \caption{Llama3.2-3B}
    \end{subfigure}
    \caption{Training curves on GSM8K. Counterfactual weighting (red) consistently outperforms vanilla DAPO (blue) throughout training. Inverted weighting (purple) underperforms, validating that importance direction matters. Shaded regions show $\pm 1$ std across seeds.}
    \label{fig:curves}
\end{figure}

Results are summarized in \cref{tab:main,tab:auc,tab:significance} and \cref{fig:curves}. We observe a consistent pattern across all three model families:

\paragraph{(1) Counterfactual weighting improves over baseline.} Counterfactual importance weighting outperforms vanilla DAPO on all models: $+0.9$ pp on Qwen3-1.7B, $+1.1$ pp on Qwen2.5-3B, and $+0.8$ pp on Llama3.2-3B. The improvement is statistically significant for Qwen2.5-3B ($p=0.002$) at step 500; AUC improvements are significant for both Qwen models ($p<0.05$). The effect is consistent throughout training (\cref{fig:curves}), not just at convergence.

\paragraph{(2) Inverted weighting underperforms.}
Inverting the importance signal - upweighting tokens that don't matter and downweighting those that do - consistently yields worse performance. On Qwen3-1.7B, inverted underperforms DAPO by $1.5$ pp; on Llama3.2-3B the gap widens to $1.8$ pp. This directional validation confirms the importance signal captures genuine causal structure rather than acting as arbitrary regularization.

\paragraph{(3) Random weighting is broadly neutral.}
Random weights perform comparably to or slightly below vanilla baseline, confirming that non-uniform weighting alone is insufficient - the \emph{direction} of importance matters. Counterfactual weighting, which assigns high weights to causally important spans, yields consistent gains across all seeds and models.

\paragraph{(4) Results hold across architectures.} Counterfactual weighting consistently outperforms all baselines across both Qwen and Llama model families, suggesting the method generalizes beyond a single architecture.

\paragraph{(5) Improved sample efficiency.}
Beyond final accuracy, counterfactual weighting accelerates learning throughout training.
On average, CF reaches DAPO's final accuracy (at step 500) substantially earlier:
at step 100 for Qwen3-1.7B,
step 125 for Qwen2.5-3B,
and step 225 for Llama3.2-3B.
This improved sample efficiency can offset the 32--74\% per-step overhead:
to reach a given target accuracy, CF often requires comparable or less total wall-clock time than vanilla DAPO.

\subsection{Analysis}

We analyze counterfactual importance scores across 13,737 spans from 1,467 completions. Beyond validating our method, this analysis reveals structural properties of reasoning traces---what content is critical, what is scaffolding, and what is actively harmful---that may inform credit assignment methods beyond our specific approach.

\paragraph{Importance distribution.}
The distribution of importance drops is heavily left-skewed (skewness $= -2.19$, kurtosis $= 7.73$), indicating that most spans have moderate importance while a minority are critical. \cref{tab:importance_distribution} shows the breakdown: only 10.9\% of spans are ``critical'' (drop $< -500$), while 45.4\% fall in the ``moderate'' range. Notably, 3.5\% of spans (481 total) are distractors: masking them actually improves answer probability, suggesting they add noise rather than signal.

\begin{table}[H]
\centering
\small
\caption{Distribution of importance drops across 13,737 spans. More negative indicates higher importance.}
\label{tab:importance_distribution}
\begin{tabular}{lrrr}
\toprule
\textbf{Category} & \textbf{Drop Range} & \textbf{Count} & \textbf{\%} \\
\midrule
Critical & $< -500$ & 1,500 & 10.9\% \\
Important & $[-500, -200)$ & 4,176 & 30.4\% \\
Moderate & $[-200, -50)$ & 6,231 & 45.4\% \\
Low & $[-50, 0)$ & 1,349 & 9.8\% \\
Distractor & $\geq 0$ & 481 & 3.5\% \\
\bottomrule
\end{tabular}
\end{table}

\paragraph{What content is critical?}
We analyze textual patterns that distinguish critical spans (drop $< -500$, N=1,500) from low-importance spans ($-50 <$ drop $< 0$, N=1,349). \cref{tab:content_patterns} reveals striking differences: \textbf{calculation chains} (spans containing sequential equations like ``$x = y = z$'') are \textbf{11.2$\times$} more prevalent in critical spans than low-importance ones. Multiplication and division operations show \textbf{6.5$\times$} enrichment, while proportion and rate reasoning shows \textbf{3.9$\times$} enrichment. Conversely, equation \emph{setup} (``let $x$ denote...'') is \textbf{depleted} in critical spans (0.37$\times$) - such spans are necessary scaffolding but not causally important for reaching the answer.

\begin{table}[H]
\centering
\small
\caption{Content pattern prevalence in critical vs.\ low-importance spans. Enrichment $>1$ indicates the pattern is more common in critical spans; $<1$ indicates depletion.}
\label{tab:content_patterns}
\begin{tabular}{lccc}
\toprule
\textbf{Pattern} & \textbf{Critical} & \textbf{Low} & \textbf{Enrich.} \\
\midrule
Calculation chain ($=\cdots=$) & 28.3\% & 2.5\% & \textbf{11.2$\times$} \\
Multiply/divide operations & 41.9\% & 6.4\% & \textbf{6.5$\times$} \\
Proportion/rate reasoning & 22.1\% & 5.6\% & \textbf{3.9$\times$} \\
Total/sum operations & 48.5\% & 17.0\% & 2.9$\times$ \\
Conclusion (``therefore...'') & 16.8\% & 7.1\% & 2.4$\times$ \\
Equation setup (``let...'') & 3.7\% & 10.1\% & 0.37$\times$ \\
Step headers (``Step 1:'') & 1.7\% & 2.8\% & 0.62$\times$ \\
\bottomrule
\end{tabular}
\end{table}

\paragraph{Span position matters.}
We partition spans into early (first third), middle, and late (last third) positions within each completion. Middle spans are most important (mean drop $= -270$), followed by late ($-233$) and early ($-201$). All pairwise differences are highly significant ($p < 10^{-10}$). This aligns with typical solution structure: early spans restate the problem, middle spans contain core derivations, and late spans state conclusions.

\paragraph{Span length correlates with importance.}
Longer spans receive higher importance scores ($r = -0.56$, $p < 10^{-100}$). This is expected: longer spans contain more reasoning content and their removal causes greater disruption. However, the moderate correlation indicates length alone does not determine importance - a verbose restatement can be long but unimportant.

\paragraph{Correct vs.\ incorrect completions.}
Importance distributions differ slightly between correct (N=2,353 spans, mean $= -221$) and incorrect (N=11,384 spans, mean $= -235$) completions, with incorrect completions showing marginally higher importance scores ($p = 0.012$). However, this small difference (6\%) suggests our method primarily measures \emph{structural} importance - which spans are causally connected to the answer - rather than \emph{correctness}. A span can be critically important for producing a wrong answer.

\paragraph{Distractor spans.}
The 481 spans with positive importance drops (3.5\%) represent content that \emph{hurts} answer probability when present. Of these, 76.1\% occur in incorrect completions, suggesting verbose or redundant reasoning correlates with failure. Manual inspection reveals distractors are typically step headers (``Step 1: Calculate the total miles...''), redundant restatements of given information (``Humans have 2 legs''), or verbose setup without computational content. The strongest distractor (drop $= +633$) is the span ``Step 1: Calculate the total miles Jerome plans to ride in the first 12 days'' - pure scaffolding that the model navigates better without.

\paragraph{Token-level weight distribution.}
Across 858,798 non-padding tokens, the weight range $[0.5, 4.0]$ is fully utilized: 1.2\% of tokens receive minimum weight (0.5$\times$) while 21.8\% receive maximum weight (4.0$\times$, including answer tokens). This confirms the method produces substantial differentiation rather than near-uniform weights.

\paragraph{Gradient concentration.}
We quantify how counterfactual weighting redistributes gradient mass compared to uniform weighting.
Across 858,798 tokens, we categorize by normalized importance: high ($\hat{I} > 0.8$), medium ($0.5 < \hat{I} \leq 0.8$), and low ($\hat{I} \leq 0.5$).
Under uniform weighting, gradient contribution is proportional to token count.
Under CF weighting, high-importance tokens (26.3\% of tokens) receive 42.5\% of gradient mass - a \textbf{1.6$\times$ concentration}.
Conversely, low-importance tokens (53.9\% of tokens) receive only 32.2\% of gradient mass - a \textbf{0.6$\times$ dilution}.
This represents a 2.7$\times$ shift in relative gradient allocation from filler tokens to causally important reasoning steps,
empirically validating the ``gradient dilution'' intuition: CF weighting focuses parameter updates on the tokens that actually determine task success.

\paragraph{Qualitative example.}
\cref{tab:qualitative} shows importance scores for a GSM8K problem where Oliver avoids mango dishes at a buffet. The problem requires computing which dishes contain mango (via fractions and sums) then subtracting from the total.

The method clearly differentiates critical reasoning from scaffolding. The fraction calculation ``$\frac{1}{6} \times 36 = 6$'' receives highest importance ($|\Delta|=806$) because it derives a quantity not given in the problem. The sum of mango dishes ($|\Delta|=646$) aggregates this result. In contrast, restatements of given information (``Total dishes: 36'', ``Mango jelly: 1'') receive 4--5$\times$ lower importance. Notably, the final arithmetic ``$36-10+2=28$'' scores lowest ($|\Delta|=102$)---once the mango count is established, the subtraction is mechanical.

\begin{table}[H]
\centering
\small
\caption{Span importance for a GSM8K solution (Answer: 28). Problem: 36 dishes; 3 have mango salsa; $\frac{1}{6}$ have fresh mango; 1 has mango jelly. Oliver avoids mango but can pick it out of 2 fresh mango dishes. How many can he eat?}
\label{tab:qualitative}
\begin{tabular}{cp{9cm}r}
\toprule
\textbf{Step} & \textbf{Reasoning Span} & \textbf{$|\Delta|$} \\
\midrule
-- & Break down the problem step by step & 387 \\
1 & Total number of dishes: 36 & 172 \\
2 & Dishes with mango salsa: 3 & 211 \\
3 & Fresh mangoes: $\frac{1}{6} \times 36 = 6$ dishes & \textbf{806} \\
4 & Dishes with mango jelly: 1 & 183 \\
5 & Total mango: $6 + 3 + 1 = 10$ dishes & \textbf{646} \\
6 & Oliver can pick mango out of: 2 & 148 \\
7 & Final: $36 - 10 + 2 = 28$ & 102 \\
-- & Oliver can eat 28 dishes & 190 \\
\bottomrule
\end{tabular}
\end{table}

\paragraph{Summary.}
Our analysis reveals that counterfactual importance successfully identifies reasoning structure: calculation chains and arithmetic operations are critical (11$\times$ and 6.5$\times$ enriched); setup and headers are scaffolding (0.4--0.6$\times$ depleted); and 3.5\% of spans are actively harmful distractors. The method captures causal importance rather than surface features, though length and position correlate as expected. These findings support using counterfactual importance to focus gradient updates on the spans that actually determine task success.

\subsection{Code Generation: A Negative Result}
\label{sec:code}

We evaluated counterfactual weighting on MBPP+ code generation to test whether the method generalizes beyond mathematical reasoning. \cref{tab:mbpp} shows results across training checkpoints.

\begin{table}[H]
\centering
\small
\caption{MBPP+ pass rate (\%) across training steps. Unlike math, counterfactual weighting provides no consistent benefit for code generation.}
\label{tab:mbpp}
\begin{tabular}{lccc}
\toprule
\textbf{Step} & \textbf{CF} & \textbf{Vanilla} & \textbf{$\Delta$} \\
\midrule
25  & 52.7 & 51.6 & +1.1 \\
100 & 52.4 & 51.9 & +0.5 \\
200 & 52.1 & 53.4 & -1.3 \\
250 & 51.3 & 54.2 & -2.9 \\
300 & 53.7 & 54.0 & -0.3 \\
\bottomrule
\end{tabular}
\end{table}

We find no significant improvement over baseline DAPO - at the best checkpoint, vanilla slightly outperforms CF (54.2\% vs 53.7\%). Early training shows a small CF advantage, but vanilla takes over from step 150 onward. Results are largely within noise ($\pm$1--2\%).

We hypothesize this null result stems from the \emph{distributed} nature of code correctness. In math, a single numeric answer serves as the verification target: masking a calculation chain destroys the unique path to that ``answer sink,'' producing a strong counterfactual signal. In code, correctness requires multiple structural elements - control flow, variable bindings, return statements - to be simultaneously correct. No single span's removal catastrophically damages output probability in the same way. Additionally, our arithmetic-focused span detection (equations, calculations) fails to capture code-relevant structure such as conditionals, loops, and function calls.

This negative result clarifies the method's scope: counterfactual importance weighting is effective when task success depends on a small number of critical reasoning steps with a concentrated verification signal, as in mathematical problem-solving.

\section{Limitations}
\label{sec:limitations}

\paragraph{Computational overhead.}
The method requires additional forward passes to estimate span importance, adding 32--74\% overhead depending on the number of spans per completion. While the optimizations in \cref{sec:overhead} (batched masking, caching, early termination) mitigate this cost, the overhead remains substantial for large-scale training. The improved sample efficiency can offset this cost, but practitioners must weigh per-step overhead against convergence benefits.

\paragraph{Domain specificity.}
Our span detection relies on regex patterns tuned for arithmetic reasoning: equations, calculations, and sentence boundaries. This works well for GSM8K but fails to capture domain-relevant structure in other settings. The null results on MBPP+ (\cref{sec:code}) demonstrate this limitation---conditionals, loops, and function definitions are not detected as spans. Extending the method to new domains requires designing appropriate span detectors, which may require domain expertise.

\paragraph{Imperfect causal intervention.}
Span masking is an approximation to true counterfactual reasoning. Replacing tokens with placeholders may introduce distributional shift: the model has never seen padding tokens mid-sequence during pretraining, so its behavior under masking may not reflect ``what would happen if this information were absent.'' Alternative interventions (e.g., resampling from the model's own distribution, or activation patching) could provide cleaner causal estimates but at higher computational cost.

\paragraph{Answer-centric importance.}
Our importance metric measures contribution to the \emph{final answer probability}. This captures causal importance for outcome but may miss spans that improve reasoning quality without directly affecting the answer (e.g., clear explanations, error checking). For tasks where intermediate reasoning quality matters beyond final accuracy, alternative importance definitions may be needed.

\paragraph{Limited scale.}
We evaluated on models up to 3B parameters. Whether counterfactual weighting provides similar benefits at larger scales (7B+) remains untested. Larger models may have more robust credit assignment internally, potentially reducing the benefit of explicit importance weighting.

\section{Conclusion}

We introduced counterfactual importance weighting for GRPO-like family of algorithms. The method identifies causally important reasoning spans by measuring answer probability drop under masking, then uses this signal to reweight token-level policy gradients. Experiments across three model families demonstrate consistent improvements of 0.8--1.1 percentage points over baseline DAPO, with faster convergence to equivalent accuracy. While these gains are modest, they are consistent across architectures, validated by ablations, and achieved with no auxiliary models---suggesting that counterfactual signal is both real and usable. Analysis of 13,737 spans reveals the method correctly identifies calculation chains as critical (11$\times$ enriched) while deprioritizing scaffolding tokens, concentrating 1.6$\times$ more gradient mass on high-importance reasoning steps.

Beyond the training improvements, our analysis reveals structural properties of reasoning traces that may be independently useful. The finding that 3.5\% of spans are outright distractors---content whose removal improves answer probability---suggests that standard uniform-credit training actively reinforces harmful content. The 11$\times$ enrichment of calculation chains in critical spans, versus 0.37$\times$ depletion of equation setup, quantifies the intuition that ``not all tokens matter equally'' and provides a basis for future work on adaptive curricula or data filtering.

The approach requires no auxiliary models, no external annotation, and adds only forward-pass overhead - providing a non-parametric alternative to Process Reward Models that extracts credit assignment signal directly from the policy model's own probability estimates. The key validation is directional: inverting the importance signal hurts performance, confirming that the method captures genuine causal structure rather than acting as arbitrary regularization.

However, the method does not transfer to code generation (\cref{sec:code}), where correctness is distributed across multiple structural elements rather than concentrated in a single answer. Future work should explore domain-specific span detection for code (e.g., control flow, function boundaries) and the computational optimizations discussed in \cref{sec:overhead}.

\section*{Impact Statement}

This paper presents work whose goal is to advance the field of Machine Learning by improving credit assignment in reinforcement learning for language models.

\paragraph{Potential benefits.}
Our method improves training efficiency, potentially reducing the computational resources and energy required to achieve a given capability level. The analysis of reasoning traces---identifying which content helps versus hurts answer probability---may contribute to more interpretable training processes. Understanding what makes reasoning steps causally important could inform curriculum design, data filtering, and debugging of model failures.

\paragraph{Potential risks.}
As with any work that improves language model training, our method could accelerate the development of more capable models, which carries dual-use concerns. However, we note that our improvements are modest (0.8--1.1 pp) and focused on mathematical reasoning rather than general capabilities. The method's reliance on verifiable answers limits its applicability to domains with clear correctness criteria, reducing concerns about training models on subjective or potentially harmful content.

\paragraph{Broader considerations.}
The finding that 3.5\% of reasoning spans are ``distractors'' whose removal improves performance suggests that current training methods may inadvertently reinforce unhelpful or misleading content. While we use this observation to improve training, it also highlights a potential failure mode of uniform credit assignment that the research community should be aware of. We do not foresee specific negative societal consequences beyond those generally associated with improved language model capabilities.

\bibliographystyle{unsrtnat}
\bibliography{references}

\appendix
\section{Computational Overhead Analysis}
\label{sec:overhead}

Our counterfactual importance method introduces computational overhead from the additional forward passes required to estimate token importance. We provide a detailed analysis of this overhead and discuss both current optimizations and potential improvements.

\paragraph{Sources of overhead.}
For each completion in a training batch, we must compute the answer probability drop under masking for every reasoning span. This requires one additional forward pass per span. \cref{tab:overhead} summarizes the overhead across our experimental configurations.

\begin{table}[H]
\centering
\small
\caption{Computational overhead of counterfactual importance estimation. Runtime measured on a single L40S GPU. Spans/completion reflects spans passing importance thresholds during training.}
\label{tab:overhead}
\begin{tabular}{lccccc}
\toprule
\textbf{Model} & \textbf{DAPO} & \textbf{CF} & \textbf{OH} & \textbf{Spans/} & \textbf{CF/} \\
 & \textbf{(min)} & \textbf{(min)} & & \textbf{Comp.} & \textbf{Step} \\
\midrule
Qwen3-1.7B & 405 & 535 & +32\% & 5.2 & 21.5s \\
Llama3.2-3B & 403 & 578 & +43\% & 7.2 & 21.0s \\
Qwen2.5-3B & 412 & 715 & +74\% & 5.8 & 36.4s \\
\bottomrule
\end{tabular}
\end{table}

The overhead varies significantly across models (32--74\%), primarily driven by the cost of forward passes on each architecture rather than the number of spans detected. Counterfactual estimation consumes 30--42\% of total training time, comparable to the generation phase itself. For Qwen2.5-3B, time per step breaks down as: generation (42.4s, 49\%), log-probs (1.2s, 1\%), CF estimation (36.4s, 42\%), and other operations (5.9s, 7\%).

\paragraph{Memory overhead.}
Interestingly, counterfactual estimation does not increase GPU memory usage. Since CF forward passes only compute log-probabilities without gradient tracking, peak memory remains dominated by the backward pass of the main training loop. We observe slightly \textit{lower} memory usage in CF runs (11.3--12.5 GB) compared to vanilla (12.6--13.3 GB), likely due to more aggressive memory clearing between CF passes.

\paragraph{Current optimizations.}
Our implementation includes several optimizations:

\begin{itemize}
\item \textbf{Gradient-free forward passes}: CF estimation uses \texttt{torch.no\_grad()}, reducing per-pass cost to approximately 0.4--0.7$\times$ a standard forward pass (0.37s vs 0.75s for Llama3.2-3B).
\item \textbf{Hash-based caching}: We cache importance weights keyed by (prompt, completion) hash to avoid recomputation for identical samples across epochs. However, due to on-policy sampling generating new completions each step, cache hit rates remain below 1\% in practice.
\item \textbf{Batched span detection}: Reasoning spans are detected via vectorized regex matching over the full batch before any forward passes, minimizing Python overhead.
\end{itemize}

\paragraph{Potential improvements.}
Several optimizations could substantially reduce overhead in future work:

\begin{enumerate}
\item \textbf{Batched CF forward passes}: Currently, masked sequences are processed with variable-length inputs. Padding and batching CF passes could improve GPU utilization by 2--3$\times$.

\item \textbf{Proxy model estimation}: Using a smaller model (e.g., Qwen-0.5B) to estimate importance weights for a larger model being trained. Initial experiments suggest importance rankings transfer reasonably across model scales.

\item \textbf{Sparse span selection}: Rather than evaluating all spans, select only the top-$k$ longest or most syntactically complex spans. With $k=3$, this could reduce CF passes by 40--60\% with minimal signal loss.

\item \textbf{Importance weight caching across epochs}: For offline RL settings where the dataset is fixed, importance weights could be precomputed once and reused, amortizing the overhead to near-zero for subsequent epochs.

\item \textbf{Asynchronous computation}: CF estimation for batch $t$ could overlap with gradient computation for batch $t-1$ using separate CUDA streams, hiding latency on multi-GPU setups.

\item \textbf{Early termination}: Skip CF computation for completions where all candidate answers receive zero reward (uniformly incorrect), as their loss contribution is already zero in DAPO.
\end{enumerate}

\paragraph{Cost-benefit analysis.}
Despite the 32--74\% overhead, counterfactual weighting improves learning efficiency as measured by accuracy gain per gradient step. The practical question is whether CF's per-step gains justify its slower iterations:

\begin{itemize}
\item \textbf{Fixed step budget}: For 500 training steps, CF takes 32--74\% longer but achieves 0.8--1.1pp higher final accuracy - a favorable tradeoff when final performance matters more than iteration speed.
\item \textbf{Fixed compute budget}: Given the same wall-clock time, vanilla DAPO completes more steps but CF still matches or exceeds its accuracy due to more efficient credit assignment per step.
\end{itemize}

We recommend vanilla DAPO for rapid hyperparameter search, with CF weighting applied for final training runs where accuracy improvements justify additional compute.

\end{document}